\documentclass[conference]{IEEEtran}
\usepackage[pdftex]{graphicx}
\usepackage[cmex10]{amsmath}

\newcommand{\op}[1]{\texttt{\textbf{#1}}}

\newcommand{\f}[1]{\textcircled{\footnotesize{#1}}}

\begin{document}
\title{A parameterized activation function for learning\\
fuzzy logic operations in deep neural networks}

\author{
	\IEEEauthorblockN{Luke B. Godfrey}
	\IEEEauthorblockA{
		Department of Computer Science and Computer Engineering\\
		University of Arkansas\\
		Fayetteville, AR 72701\\
	}
	\and
	\IEEEauthorblockN{Michael S. Gashler}
	\IEEEauthorblockA{
		Department of Computer Science and Computer Engineering\\
		University of Arkansas\\
		Fayetteville, AR 72701\\
	}
}

\maketitle

\begin{abstract}
We present a deep learning architecture for learning fuzzy logic expressions.
Our model uses an innovative, parameterized, differentiable activation function that can learn a number of logical operations by gradient descent.
This activation function allows a neural network to determine the relationships between its input variables and provides insight into the logical significance of learned network parameters.
We provide a theoretical basis for this parameterization and demonstrate its effectiveness and utility by successfully applying our model to five classification problems from the UCI Machine Learning Repository.
\end{abstract}

\IEEEpeerreviewmaketitle

\section{Introduction} \label{sec_introduction}

Neural networks are powerful adaptive models with applications in many disciplines.
With the backpropagation algorithm, neural networks can be trained in a straightforward manner by gradient-based optimization techniques like stochastic gradient descent and RMSProp \cite{tieleman2012lecture}.
Some of these models, such as convolutional neural networks, learn parameters that can be visualized, interpreted, and understood in some cases \cite{zeiler2014visualizing}.
Most neural networks, however, are considered black boxes \cite{kanungo2006comparative,sjoberg1995nonlinear} and it is difficult to determine the semantic meanings behind learned weights.

Fuzzy inference systems, built on fuzzy logic, are also powerful models.
Unlike neural networks, fuzzy inference systems are straightforward to interpret and often use linguistic values \cite{zadeh1996fuzzy}.
In fact, these systems are functionally equivalent to a subset of neural networks \cite{jang1993functional}.
Fuzzy inference systems are less general than neural networks, however, and many neural network techniques are not easily translated into the domain of fuzzy logic.

For these reasons, there is great interest in combining neural networks with fuzzy logic and fuzzy inference systems.
Adaptive fuzzy systems have been studied for decades \cite{jang1991fuzzy,jang1993anfis,lin1996neural}, and neural fuzzy modeling continues to be an active topic of research \cite{chen2013solar,chen2014fuzzy,kayacan2015identification}.
One purpose of combining these techniques is to produce a model with the flexibility and accuracy of black-box neural networks and the interpretability of fuzzy systems.

Although neural fuzzy systems are well-studied, existing approaches primarily focus on combining specific logical operations in a predefined manner, the most common being an \op{or} of \op{and}s \cite{jang1991fuzzy,jang1993anfis,kwan1994fuzzy}.
Models that restrict themselves to particular kinds of expressions are limited in the insights they can offer about given datasets.
A system that can adaptively choose from a larger set of logical operations, on the other hand, would be able to provide us with more knowledge about the relationships between its various inputs.

We present a deep learning architecture for learning fuzzy logic expressions by using a novel adaptive transfer function.
Our model uses an innovative, parameterized, differentiable activation function that can learn a number of logical operations by gradient descent.
Parameters learned by our model can be interpreted as fuzzy rules and combined to form complex logic expressions, allowing a glimpse into the knowledge gleaned during the training process.
In Section~\ref{sec_validation}, we report the results of applying our model to five classification problems taken from the UCI Machine Learning Repository \cite{uci:2007}.
We find that our model is able to learn complex logical expressions and to achieve accuracy comparable to a standard deep neural network with \op{tanh} activation functions.

\section{Fuzzy Logic and Neural Fuzzy Systems} \label{sec_related}

Fuzzy logic \cite{klir1995fuzzy} extends boolean logic into a continuous domain.
Typically, \op{false} is represented as \op{0}, \op{true} is represented as \op{1}, and values in between indicate a corresponding ``fuzzy'' degree of uncertainty.
A typical set of fuzzy operators that generalize the behavior of boolean logic are:

\begin{center}
	\begin{tabular}{ l l }
		$\op{identity}(x)$ & $=x$ \\
		$\op{not}(x)$ & $=1-x$ \\
		$\op{or}(x,y)$ & $=1-(1-x)\cdot (1-y)$ \\
		$\op{xor}(x,y)$ & $=x+y-2\cdot x \cdot y$\\
		$\op{and}(x,y)$ & $=x\cdot y$ \\
		$\op{nor}(x,y)$ & $=(1-x)\cdot (1-y)$ \\
		$\op{nxor}(x,y)$ & $=1-(x+y-2\cdot x \cdot y)$\\
		$\op{nand}(x,y)$ & $=1-x\cdot y$ \\
	\end{tabular}
\end{center}

One of the most common applications of fuzzy logic is to control systems \cite{lee1990fuzzy} using a fuzzy inference system \cite{zadeh1988fuzzy}.
A typical fuzzy logic controller uses Gaussian membership functions to ``fuzzify'' inputs (which may be linguistic \cite{zadeh1996fuzzy}), a set of (fuzzy) logical IF-THEN rules to apply to the fuzzified inputs, and a function that aggregates and ``defuzzifies'' the result to a crisp value that determines the system's action \cite{jang1993functional}.
For example, a fuzzy inference system for an autonomous car might have this inference rule: \textit{IF speed limit IS low OR traffic IS dense THEN speed = slow}.
In this example, \textit{low}, \textit{dense}, and \textit{slow} are linguistic values that correspond to functions that map raw sensor inputs to real numbers in the range $[0..1]$ where $0$ indicates definitely not in the given set, $1$ indicates definitely in the set, and anything else represents how typical the input is for the given set.
100 km/h might be 0.5 in the moderate speed set and 0.8 in the fast speed set.

The combination of fuzzy logic with neural networks has been termed ``fuzzy modeling'' \cite{jang1991fuzzy}, ``neural fuzzy systems'' \cite{lin1996neural}, and ``adaptive fuzzy systems'' \cite{jang1993anfis}.
These systems were extensively studied in the 1990s \cite{kosko1992neural,chen1995model,hayashi1993fuzzy,carpenter1992fuzzy,zhenya1998extracting,jang1997neuro}, and one of the primary reasons was that the weights and parameters of a neural fuzzy system could be interpreted more meaningfully than in a traditional neural network \cite{cox1993adaptive}.
More recently, fuzzy neural networks have been applied to tracking control \cite{chen2014fuzzy} and other nonlinear dynamics \cite{kayacan2015identification}.

One kind of neural fuzzy system is any neural network that directly models a fuzzy inference system \cite{lin1991neural,lin1994reinforcement,chen1995model}.
These models generally have five layers: 1) an input layer, 2) a membership layer to fuzzify input, 3) a rules layer that computes products of the second layers outputs, 4) a normalization layer to defuzzify signals, and 5) a summation layer to produce the output.
Inputs and outputs to this kind of system are crisp, and the fuzzy logic takes place in the hidden layers of the network.
In 1993, Jang and Sun showed that these neural networks are functionally equivalent to fuzzy inference systems \cite{jang1993functional}.
The learning that occurs in this kind of neural fuzzy system tunes the member functions, adjusting means and standard deviations, in addition to the combination weights in the output layer.
The rules layer in these models is implemented as a product-of-inputs \cite{jang1991fuzzy}, which is a logical \op{and}.
The summation layer performs the logical \op{or}, and so most of these models result in a logical \op{or} of \op{and}s (or a \op{max} of \op{min}s \cite{kwan1994fuzzy}).

Another type of neural fuzzy system is any system that uses both fuzzy logic and neural networks as parts of a whole.
Chen et. al recently applied this kind of model to solar radiation forecasting, in which the authors used a fuzzy inference system to combine the predictions three separate neural networks like a weighted ensemble \cite{chen2013solar}.
Other models combine fuzzy systems with genetic algorithms \cite{valdez2011improved} or with a Kalman filter \cite{jang1991fuzzy}.
Still others allow inputs, weights, and outputs to be fuzzy \cite{kuo2001intelligent}.
Kwan and Cai proposed the use of ``fuzzy neurons'' that combine an aggregation function and an activation function with some number of membership functions \cite{kwan1994fuzzy}.

\section{Approach}

We take an approach similar to the common five-layer neural fuzzy system \cite{jang1993anfis}, although we use a deeper network.
Our model is unique not in its topology in the interpretability of its weights, however, but in the adaptive activation function we use that is able to learn several logical operations.
Our activation function is parameterized, continuous, and differentiable, and can therefore be tuned by gradient descent.
This has an advantage over existing neural fuzzy systems because it can model more than just an \op{or} of \op{and}s, and it has an advantage over other fuzzy neuron approaches because it can be trained by gradient descent instead of set by hand.

The fuzzy logic operators listed in Section~\ref{sec_related} are elegant because they are simple and continuous.
However, the symmetry in these equations is difficult to see because our values are not centered about the origin.
If we linearly remap these operations by defining \op{false} to be \op{-1} instead of \op{0}, they become:

\begin{center}
	\begin{tabular}{ l l }
		$\op{identity}(x)$ & $=x$ \\
		$\op{not}(x)$ & $=-x$ \\
		$\op{or}(x,y)$ & $=-\frac{(x-1)(y-1)}{2}+1$ \\
		$\op{xor}(x,y)$ & $=-x\cdot y$\\
		$\op{and}(x,y)$ & $=\frac{(x+1)(y+1)}{2}-1$ \\
		$\op{nor}(x,y)$ & $=\frac{(x-1)(y-1)}{2}-1$ \\
		$\op{nxor}(x,y)$ & $=x\cdot y$\\
		$\op{nand}(x,y)$ & $=-\frac{(x+1)(y+1)}{2}+1$ \\
	\end{tabular}
\end{center}

Then, if we rewrite them in a consistent form, we obtain:

\begin{center}
	\begin{tabular}{ l l }
		$\op{identity}(x)$ & $=+\left(\frac{(x+0)(1+0)}{1}-0\right)$ \\
		$\op{not}(x)$ & $=+\left(\frac{(x+0)(-1+0)}{1}-0\right)$ \\
		$\op{or}(x,y)$ & $=-\left(\frac{(x-1)(y-1)}{2}-1\right)$ \\
		$\op{xor}(x,y)$ & $=-\left(\frac{(x+0)(y+0)}{1}-0\right)$\\
		$\op{and}(x,y)$ & $=+\left(\frac{(x+1)(y+1)}{2}-1\right)$ \\
		$\op{nor}(x,y)$ & $=+\left(\frac{(x-1)(y-1)}{2}-1\right)$ \\
		$\op{nxor}(x,y)$ & $=+\left(\frac{(x+0)(y+0)}{1}-0\right)$\\
		$\op{nand}(x,y)$ & $=-\left(\frac{(x+1)(y+1)}{2}-1\right)$ \\
	\end{tabular}
\end{center}

In this form, it is much more apparent that there is symmetry that can be leveraged to unify these operations into a single more general operation.
There are many possible functions that perfectly express all of these fuzzy logic operations.
Three representative solutions are given in Equations~\ref{eq_doesnotwork1}, \ref{eq_works}, and \ref{eq_doesnotwork2}.

\begin{equation}
	x\mbox{\f{$\alpha$}}y = \frac{(x+\alpha)(y+\alpha)}{\alpha^2+1}-\alpha^2\\
	\label{eq_doesnotwork1}
\end{equation}

\begin{equation}
	x\mbox{\f{$\alpha$}}y = \frac{(x+\alpha)(y+\alpha)}{|\alpha|+1}-|\alpha|\\
 	\label{eq_works}
\end{equation}

\begin{equation}
	\begin{array}{l}
	x\mbox{\f{$\alpha$}}y = \frac{t}{|t|}\sqrt{|t|}-|\alpha|\\
	\mbox{where } t = (x+\alpha)(y+\alpha)\\
	\end{array}
	\label{eq_doesnotwork2}
\end{equation}

We can plug any of these functions into our table of logical operations to obtain:

\begin{center}
	\begin{tabular}{ l l l }
		$\op{identity}(x)$ & $=x$ & $=$\op{ true}\f{0}$x$\\
		$\op{not}(x)$ & $=-x$ & $=$\op{ false}\f{0}$x$\\
		$\op{or}(x,y)$ & $=$\op{not}$(x$\f{-1}$y)$ & $=$\op{ false}\f{0}$(x$\f{-1}$y)$\\
		$\op{xor}(x,y)$ & $=$\op{not}$(x$\f{0}$y)$ & $=$\op{ false}\f{0}$(x$\f{0}$y)$\\
		$\op{and}(x,y)$ & $=x$\f{1}$y$\\
		$\op{nor}(x,y)$ & $=x$\f{-1}$y$\\
		$\op{nxor}(x,y)$ & $=x$\f{0}$y$\\
		$\op{nand}(x,y)$ & $=$\op{not}$(x$\f{1}$y)$ & $=$\op{ false}\f{0}$(x$\f{1}$y)$\\
	\end{tabular}
\end{center}

\begin{figure*}
	\label{fig_plot}
	\begin{center}
		\includegraphics[width=6.5in]{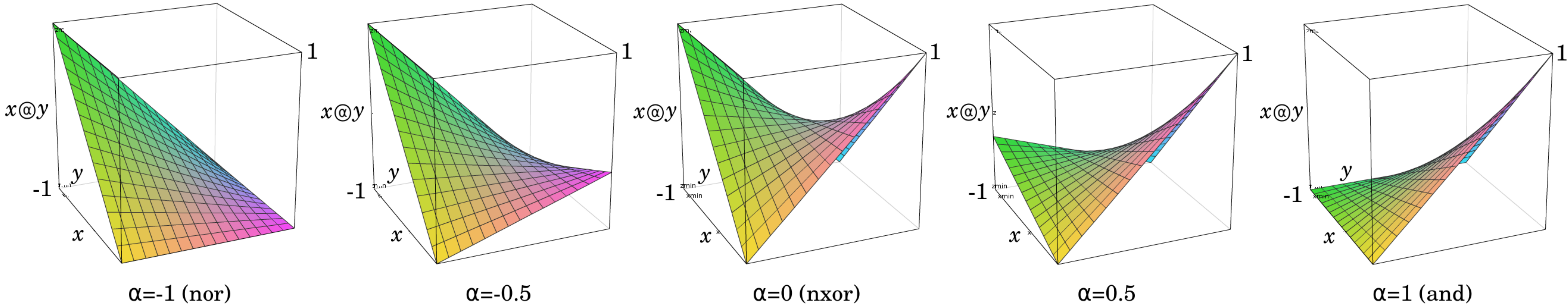}
		\caption{
			Equation~\ref{eq_works} continuously interpolates among three fuzzy logic operations: \op{nor}, \op{nxor}, and \op{and}.
			By allowing biases (\op{true} and \op{false}) and weights (in particular, a weight of -1), this equation can also compute \op{identity}, \op{not}, \op{or}, \op{xor}, and \op{nand}.
		}
	\end{center}
\end{figure*}

Figure~\ref{fig_plot} shows a plot of Equation~\ref{eq_works} with 5 values for $\alpha$.

\subsection{Learning Simple Logic Operations}\label{sec_simple}

Since Equations~\ref{eq_doesnotwork1}, \ref{eq_works}, and \ref{eq_doesnotwork2} are continuous and differentiable,
gradient-based optimization techniques could potentially be used with them to find the values for $\alpha$ that approximate the logic represented in a set of training examples.

An important and consideration in using gradient descent with fuzzy logic that does not typically occur in more traditional applications for gradient descent is that each training example only provides information about a subset of the parameter space.
For example, suppose $\alpha$ is initialized to a random value between $-1$ and $1$, and suppose the training pattern $1$\f{$\alpha$}$1=1$ is presented for optimizing the value of $\alpha$ by gradient descent.
This training pattern suggests that $\alpha$ should not be less than 0, because $(1$~\op{nor}~$1) \neq 1$.
However, this training pattern does not suggest anything about what specific value $\alpha$ should take $\ge 0$, because $(1$~\op{nxor}~$1)=1$ and $(1$~\op{and}~$1)=1$ are both equally true.

\begin{figure}[!tb]
	\begin{center}
		\includegraphics[width=3.2in]{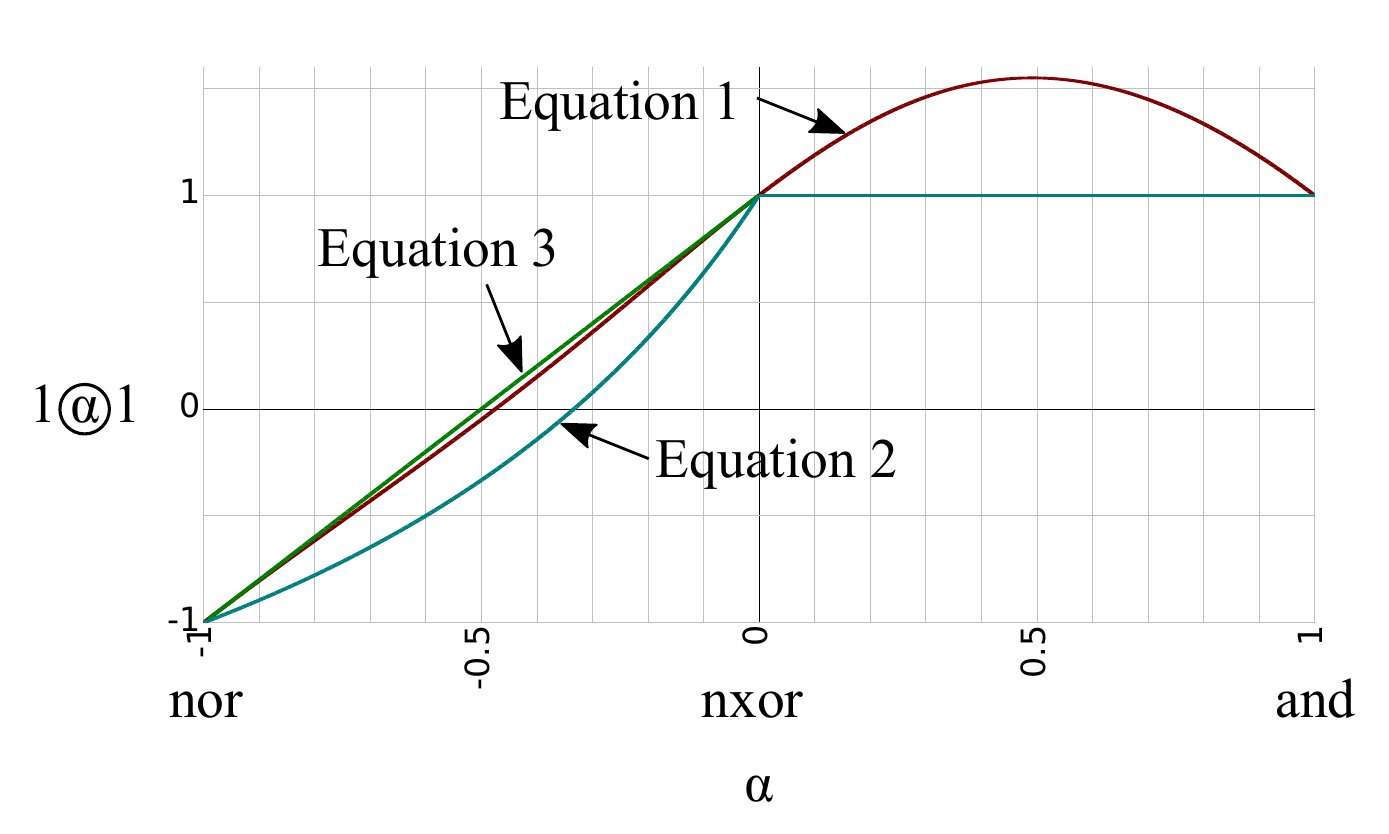}
		\caption{A comparison of Equations~\ref{eq_doesnotwork1}, \ref{eq_works}, and \ref{eq_doesnotwork2} when computing \op{true}\f{$\alpha$}\op{true}.
			Equation~\ref{eq_doesnotwork1} is smoother, but Equation~\ref{eq_works} is better suited for use with gradient-based optimization.}
	\end{center}
	\label{fig_tt}
\end{figure}

Figure~\ref{fig_tt} shows a comparison of Equations~\ref{eq_doesnotwork1}, \ref{eq_works}, and \ref{eq_doesnotwork2} for the case of computing $1$\f{$\alpha$}$1$.
If $\alpha$ has a value less than 0, then gradient-based optimization methods will adjust $\alpha$ by moving it closer to $0$ no matter which of these three equations is used.
As long as the curve in this region is monotonic, the precise shape is not important because these values are not defined in boolean logic.
However, if Equation~\ref{eq_doesnotwork1} is used, and $\alpha$ has a value greater than 0, then gradient-based optimization methods will move $\alpha$ closer to $0$ or $1$, whichever is closer to the current value of $\alpha$.
This is incorrect behavior because this training pattern does not provide any information about whether the logical operation should be more like \op{nxor} or more like \op{and}.
Since both of these operations are consistent with the training pattern, it would be arbitrary to bias the model in favor of one over the other.
Arbitrary parameter adjustments are likely to fight against subsequent training pattern presentations that may convey valid information for that region of the parameter space, resulting in the model getting stuck in a local optimum.
Equations~\ref{eq_works} and \ref{eq_doesnotwork2} correctly adjust $\alpha$ in all regions of the parameter space with this training pattern.

\begin{figure}[!tb]
	\begin{center}
		\includegraphics[width=3.2in]{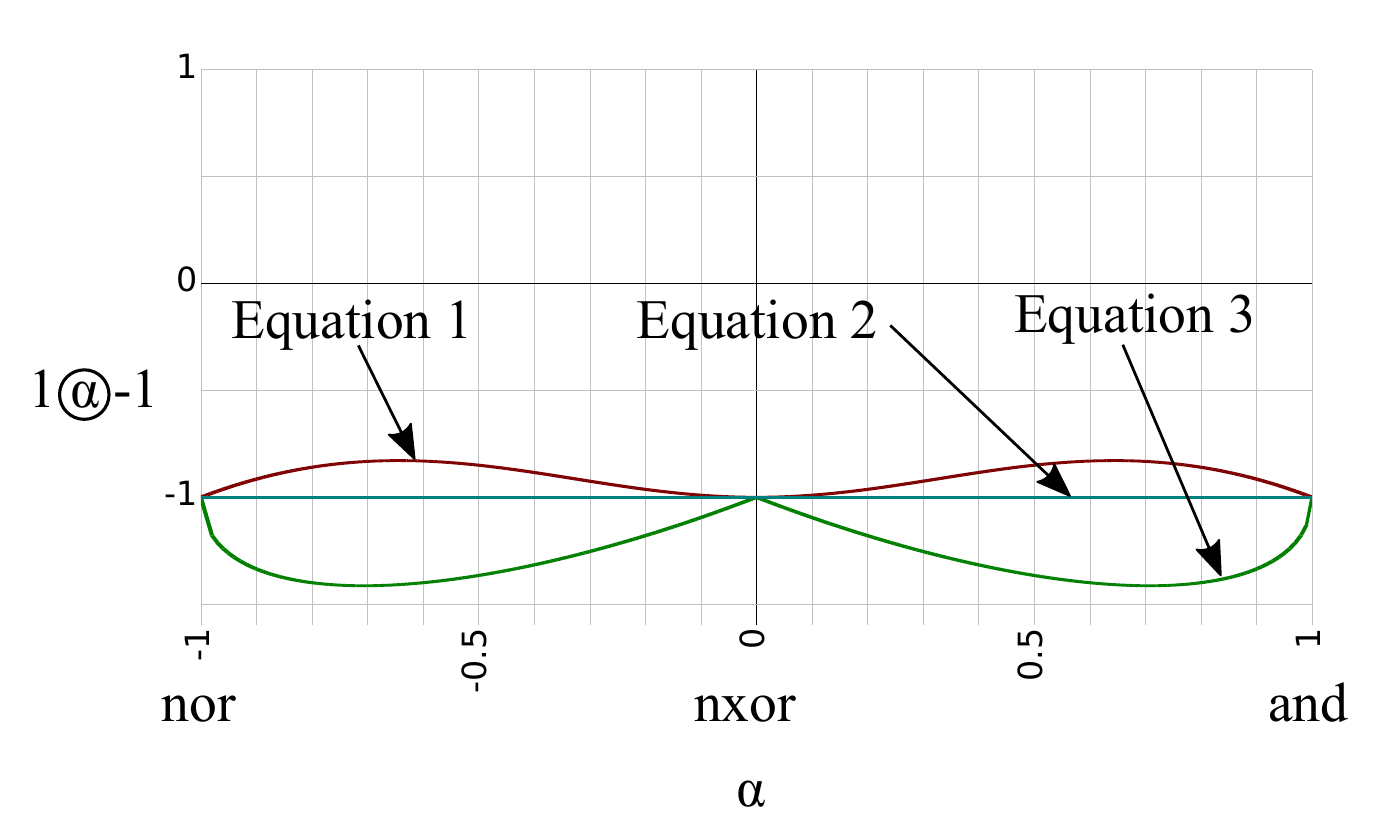}
		\caption{A comparison of Equations~\ref{eq_doesnotwork1}, \ref{eq_works}, and \ref{eq_doesnotwork2} when computing \op{true}\f{$\alpha$}\op{false}. Only Equation~\ref{eq_works} is well-suited for use with gradient-based optimization.}
	\end{center}
	\label{fig_tf}
\end{figure}

As another representative case, consider the training pattern $1$\f{$\alpha$}$-1=1$.
Perhaps counterintuitively, this pattern provides no information about any region of the parameter space, because $1$\op{nor}$-1$, $1$\op{nxor}$-1$, and $1$\op{and}$-1$ all evaluate to $-1$.
Correct behavior, therefore, should not adjust the value of $\alpha$, regardless of its current value.
Figure~\ref{fig_tf} shows that only Equation~\ref{eq_works} exhibits the correct behavior for this case.
(This does not imply that such patterns should be discarded because in a network containing many fuzzy logic units, different input values would reach each of the units depending on the current $\alpha$ values.)

The remaining two cases are both mirror images of these cases, so the same analysis applies.
It follows that Equation~\ref{eq_works} can be expected to yield correct behavior with gradient-based optimization in all cases when training with pure boolean logic.
Consistent with this intuition, we found experimentally that Equation~\ref{eq_works} was always able to learning simple boolean logical expressions,
while the other equations sometimes became stuck in local optima.
Therefore, we used Equation~\ref{eq_works} to implement $x$\f{$\alpha$}$y$ with the remainder of our experiments.

At this point, we must note a potential problem with our approach: Equation~\ref{eq_works} is not a t-norm.
A t-norm must be commutative, monotonic, and associative, and the value 1 must be the identity element \cite{gupta1991theory}.
Although our equation is commmutative and monotonic, it is not associative and the value 1 is not neutral for all $\alpha$.
Our work is, therefore, not t-norm fuzzy logic but belongs instead to a broader class of fuzzy logic.
For the purpose of this paper, we use a relaxed definition of fuzzy logic as being any logic with continuous values between \op{true} and \op{false}.

\subsection{Learning Complex Logic Expressions}\label{sec_complex}

We refer to a layer of network units that implement Equation~\ref{eq_works} as a \emph{Fuzzy} layer.
Thus, Equation~\ref{eq_works} is used as a sort of adaptive transfer function.
The only parameters to train in a fuzzy layer are one $\alpha$ value per each unit.
Equations~\ref{eq_dx}, \ref{eq_dy}, and \ref{eq_da} give the partial derivatives of Equation~\ref{eq_works}, which are necessary to train such layers with gradient-based optimization methods.

\begin{equation}
	\hspace{-0mm}\frac{\partial \mbox{\f{$\alpha$}}}{\partial x} = \frac{y+\alpha}{|\alpha|+1}
	\label{eq_dx}
\end{equation}

\begin{equation}
	\hspace{-0mm}\frac{\partial \mbox{\f{$\alpha$}}}{\partial y} = \frac{x+\alpha}{|\alpha|+1}
	\label{eq_dy}
\end{equation}

\begin{equation}
	\frac{\partial \mbox{\f{$\alpha$}}}{\partial \alpha} = \frac{|a|(x + y) - a(xy + 1)}{|a|(|a|+1)^2}
	\label{eq_da}
\end{equation}

Most gradient-based optimization methods can be implemented in three steps: (1) A forward propagation step that computes predictions with current values, (2) a backpropagation step that computes ``blame'' terms for each layer in the model, and (3) an update step that refines the parameters of the model. Equation~\ref{eq_works} is used in the forward propagation step. Equations~\ref{eq_dx} and \ref{eq_dy} are used in the backpropagation step to assign blame to preceding layers. Equation~\ref{eq_da} is used in the update step to refine the $\alpha$ values.

Because Equation~\ref{eq_da} is not continuous at $\alpha=0$, special care must be taken in the implementation of Fuzzy layers to ensure that they do not become stuck at this point.
We addressed this problem in our implementation by adding the statement ``\textbf{if} $\alpha < \epsilon$ \textbf{then} $\alpha \leftarrow -\alpha$'' to our update step.
This small addition enables $\alpha$ to cross over the value $0$ in cases where it would otherwise become stuck.
As long as $\epsilon$ is a small value, this will have negligible impact on training precision.
We used the value $\epsilon=0.001$.

Another challenge that arises in learning fuzzy logic is that Equation~\ref{eq_works} only accepts two input values, $x$ and $y$.
One possible solution is to try to generalize the equations in a manner that can support vectors of arbitrary dimensionality.
Equation~\ref{eq_doesnotwork2} can be generalized in this manner, as given in Equation~\ref{eq_doesnotwork2_general}.

\begin{equation}
	\begin{array}{l}
	f(\overrightarrow{x},\alpha) = \frac{t}{|t|}|t|^{1/n}-|\alpha| \vspace{8 pt}\\
	\mbox{where } t = \prod_i^n (x_i+\alpha)\\
	\end{array}
	\label{eq_doesnotwork2_general}
\end{equation}

(In Equation~\ref{eq_doesnotwork2_general}, $f$ is the fuzzy operator, and $n$ refers to the number of elements in $\overrightarrow{x}$.)
In higher dimensions, \op{or} becomes \op{any}, \op{and} becomes \op{all}, and \op{xor} becomes \op{parity}.
Unfortunately, Equation~\ref{eq_doesnotwork2} resists gradient-based optimization, so its generalized version is unlikely to do any better, and Equation~\ref{eq_works} cannot be generalized in this manner.
Further, in applications with many variables, it is often unlikely that all variables will simultaneously take the same value, which renders the \op{any} and \op{all} operations to have very limited utility.
Therefore, a good solution should provide a mechanism to select which values feed into each logical operation.
This is also consistent with most real-world uses of boolean logical expressions, and logical expressions involving only two variables at a time are more likely to be easily comprehensible to humans than logic involving many values.
Our solution to this challenge introduces two additional layer types, which we call \emph{AllPairings}, and \emph{FeatureSelector}.

An AllPairings layer accepts $n$ inputs, and outputs all $n(n-1)/2$ possible unordered pairings of its input values.
In order to facilitate the \op{identity} and \op{not} operations, we additionally pair each input value with the bias values \op{true} and \op{false}, which increases the total number of unordered output pairs to $n(n-1)/2+2n$.
This layer type contains no parameters to train, so it is straightforward to use in a deep network.
During the backpropagation step, the blame term assigned to each input unit is simply the sum of the blame terms for all affect output units.

A FeatureSelector layer is identical to a traditional fully-connected linear layer, except with four minor modifications:
(1) No bias weights are used,
(2) The weights are initialized with uniform values instead of random values,
(3) The weights that feed into each unit are constrained to have values between -1 and 1, and
(4) $L^1$ regularization is applied to the weights in this layer to gently promote sparse connections in this layer.
As it is closely related to a fully-connected layer, it produces a weighted sum of its inputs.
This allows the network to learn an interpolation between logical expressions learned in the Fuzzy layers.
The outputs of a FeatureSelector layer can be re-mapped between -1 and 1 (i.e. through a membership function) before they are used as input to other layers.

Our topology, given the definitions of these two layer types, is as follows.
Given $n$ continuous input values, we first normalize them between -1 and 1; this can be thought of as a single linear membership function (where 1 is ``high'' and -1 is ``low'').
Next, we feed the normalized values into an AllPairings layer.
We then feed all combinations of value pairings into a Fuzzy layer, which learns an optimal logical operation for each pair of values.
The output of the Fuzzy layer is fed into a FeatureSelector layer to manage dimensionality and to produce the desired number of output values.
If a deeper topology is desired, we can feed the output of the FeatureSelector layer into another normalizing layer (i.e. a single membership function for each output) and repeat the sequence of AllPairings, Fuzzy, and FeatureSelector layers to an arbitrary depth.
If no further depth is needed, we can feed the output of the final FeatureSelector layer into any kind of output layer; we use a \op{max} layer for classification.

In our validation, we fix the depth of logic layer to two, resulting in the following final 10-layer deep topology:
(1) the input layer (identity),
(2) a normalization layer (a single linear membership function),
(3) an AllPairings layer,
(4) a Fuzzy layer,
(5) a FeatureSelector layer,
(6) a \op{tanh} layer (a single nonlinear membership function with a new input space),
(7) another AllPairings layer,
(8) another Fuzzy layer,
(9) another FeatureSelector layer, and
(10) a \op{max} layer for classification.

\section{Validation}\label{sec_validation}

Our goal in this work is not to surpass the accuracy of existing results or to claim any novelty in learning fuzzy logic rules, but to demonstrate an alternative approach to fuzzy learning using a novel adaptive transfer function.
We apply our model to five classification problems taken from the UCI Machine Learning Repository \cite{uci:2007}, comparing it with a regular deep neural network (DNN) with \op{tanh} activation functions as well as with ensemble of fuzzy classifiers (EFC) proposed by Canul-Reich et. al in 2007 \cite{canul2007ensembles}.
Our results demonstrate that our model meets our goal, adaptively learning complex logic expressions by gradient descent while yielding accuracies comparable to existing methods and learning fuzzy logic rules.

\begin{table}
	\caption{
		Average errors of a deep neural network (DNN), an ensemble of ANFIS fuzzy classifiers (EFC), our model, and the expressions obtained from ``snapping'' the weights in our network (Snapped) on the validation data for five classification problems.
		Best results are \textbf{bolded}.
	}
	\begin{center}
		\begin{tabular}{ | l | l | l | l | l | }
			\hline
			Dataset			& DNN				& EFC		& Our Model			& Snapped	\\
			\hline
			Breast Cancer	& 3.26\%			& 6.42\%	& \textbf{2.77\%}	& 2.84\%	\\
			Diabetes		& 29.68\%			& 24.51\%	& \textbf{22.79\%}	& 35.06\%	\\
			Vehicle			& \textbf{18.01\%}	& 50.58\%	& 28.71\%			& 67.84\%	\\
			Waveform		& \textbf{14.95\%}	& -			& 15.27\%			& 68.43\%	\\
			Yeast			& \textbf{46.12\%}	& 67.37\%	& 49.77\%			& 82.94\%	\\
			\hline
		\end{tabular}
	\end{center}
	\label{tab_results}
\end{table}

The five classification problems we used were breast cancer, diabetes, vehicle, waveform, and yeast.
Errors yielded by four models are compared in Table~\ref{tab_results}.
Across all five problems, we fixed our models parameters to demonstrate robustness.
In particular, we set the learning rate to 0.01 and the regularization term to 0.0001.
We use a DNN topology with a depth and number of weights similar to our model, but with each layer being fully connected and all activation functions set to \op{tanh}.

We also include the results reported by Canul-Reich et. al \cite{canul2007ensembles} rather than re-evaluating their method ourselves (which is why there is no result for their model on the waveform problem).
Their model is an ANFIS-based ensemble of fuzzy classifiers (labeled EFC in Table~\ref{tab_results}).
One of the important observations made by Canul-Reich et. al was that their model performed poorly on datasets with more than six input features (vehicle and yeast).

All of the datasets we use have continuous inputs and discrete class outputs.
Breast cancer has 9 inputs and 2-class outputs. Our model outperforms both DNN and EFC in terms of accuracy.
Diabetes (Pima) has 8 inputs and 2-class outputs. Again, our model achieves higher accuracy than the compared models.

Vehicle has 18 inputs and 4-class outputs.
This is an interesting problem because it has a high number of inputs and more than a binary output.
The DNN has significantly lower error than our model, likely due to the difficulties fuzzy systems typically have with larger input spaces.
However, our model makes more accurate predictions than EFC, demonstrating that although our model is susceptible to problems with large input spaces, it is more robust than some other fuzzy-based approaches.

Waveform has 40 inputs and 3-class outputs.
This is a very large input space, so it is an interesting test for a fuzzy-based system like ours.
Our model performed well on this problem, yielding test errors only marginally higher than the DNN.
(Canul-Reich et. al did not report results on this problem for EFC).

Yeast has 8 inputs and 10-class outputs.
Out of the five datasets we used, this had the highest number of output classes, so the results of this test demonstrate how our model handles many classes.
Our model performed nearly as well as the DNN and significantly better than the EFC, demonstrating that it is able to model data with many partitions.

Our model's weights can be interpreted as complex logical expressions that describe the relationships learned between various inputs.
We form these expressions by ``snapping'' the $\alpha$ parameters to the nearest whole value (i.e. $\alpha = -0.2$ snaps to $0$ to become \op{nxor}) and  ``snapping'' the weights on the FeatureSelector layer to the nearest whole value (effectively negating expressions with weights near -1 and dropping expressions with weights near 0).
We use $n$ to denote that input $n$ is ``high'', $\neg$ for \op{not}, $\&$ for \op{and}, $|$ for \op{or}, and $\oplus$ for \op{xor}.
$c_i$ is the output for class $i$, where the $i$ with maximum $c_i$ is the predicted class.
Addition has no equivalent logical operation, so we leave it as a sum indicating the interpolation between operands.
(In the network, the sum is fed through a \op{tanh} function to map the result back into our logical space; this step is omitted in the snapped expressions).
The resulting expressions for three of the problems (breast cancer, diabetes, and vehicle) are shown in Table~\ref{tab_exp}; expressions for waveform and yeast were omitted from this paper for brevity.
Accuracies of the formed expressions applied to the validation data are reported in the ``Snapped'' column of Table~\ref{tab_results}.

\begin{table}
	\caption{
		Logical expressions learned by our model for three of the datasets.
		These expressions were formed by ``snapping'' all parameters of the network to the nearest whole value.
		Expressions for the other two datasets were omitted for the sake of brevity.
	}
	\begin{center}
		\begin{tabular}{ | l | l | }
			\hline
			Dataset			& Expression																							\\
			\hline
			Breast Cancer	& $c_0 = (0 \,|\, 3) + (1 \,|\, 5)$																		\\
							& $c_1 = ((1 \,|\, 3) + (3 \,|\, 7)) \,|\, ((2 \,|\, 8) + (3 \,|\, 4) + (5 \,|\, 6))$					\\
			\hline
			Diabetes		& $c_0 = (\neg(2 \,|\, 4) + 7 + \neg(2 \,|\, 6)) \,|\, (\neg(2 \,\&\, 6) + \neg0)$						\\
							& $c_1 = (\neg(2 + \neg(1 \,\&\, 5) + 1) \,\&\, (2 \,|\, 6))$											\\
			\hline
			Vehicle			& $\begin{aligned} c_0 ={} & \neg((0 \,\&\, 4) \,|\, ((6 \,\&\, 10) + (4 \,\&\, 16) + (7 \,\&\, 13)))	\\
										& + (\neg(2 \,|\, 16) \,\&\, ((4 \,\&\, 9) + (5 \,\&\, 6))) \end{aligned}$					\\
							& $c_1 = (4 \,\&\, 9) + (5 \,\&\, 6) + \neg(((0 \,\&\, 4)) \,\oplus\, (8 \,\&\, 16))$			\\
							& $c_2 = \neg(((3 \,\&\, 5) + (0 \,\&\, 4) + (3 \,\&\, 13)) \,\&\, (0 \,\&\, 4))$						\\
							& $c_3 = \neg((5 \,\&\, 15) \,\&\, (0 \,\&\, 4))$														\\
			\hline
		\end{tabular}
	\end{center}
	\label{tab_exp}
\end{table}

\section{Conclusion}\label{sec_conclusion}

We presented a deep learning architecture for learning fuzzy logic expressions.
The primary component of this architecture is an innovative, parameterized, differentiable activation function that can learn a number of logical operations by gradient descent.
This activation function is unique to our approach and unifies fuzzy logic with deep learning in a new way that leverages the advantages of both domains.
We provided both a theoretical basis for our activation function and testing results on five classification problems from the UCI Machine Learning Repository.
Not only was our model a reasonably strong classifier for these problems, but also the parameters of our model were interpretable as logical expressions that summarized what it had learned.
Our primary contribution is neither a better classifier nor the first model whose weights can be interpreted as fuzzy logic rules, however, but the adaptive transfer function that learns logical operations by gradient descent.
We believe that this is a promising new approach to combining fuzzy logic with deep learning that potentially opens the way for future work in both domains.

\IEEEtriggeratref{9}
\bibliographystyle{IEEEtran}
\bibliography{refs}

\end{document}